\documentclass[a4paper,fleqn]{cas-sc}

\usepackage[numbers]{natbib}
\usepackage[ruled]{algorithm2e}

\def\tsc#1{\csdef{#1}{\textsc{\lowercase{#1}}\xspace}}
\tsc{WGM}
\tsc{QE}

\begin{document}
\let\WriteBookmarks\relax
\def\floatpagepagefraction{1}
\def\textpagefraction{.001}

\shorttitle{}    

\shortauthors{}  

\title [mode = title]{Targeted Efficient Fine-tuning: Optimizing Parameter Updates with Data-Driven Sample Selection}  



%

\author[1,2,3]{Ming Dong}[
orcid=0000-0003-3700-0154,
]



\ead{dongming@ccnu.edu.cn}



\affiliation[1]{organization={Hubei Provincial Key Laboratory of Artificial Intelligence and Smart Learning},
            addressline={}, 
            city={Wuhan},
            postcode={430079}, 
            state={Hubei},
            country={PR China}}

\affiliation[2]{organization={School of Computer, Central China Normal University},
	addressline={}, 
	city={Wuhan},
	postcode={430079}, 
	state={Hubei},
	country={PR China}}

\affiliation[3]{organization={National Language Resources Monitoring and Research Center for Network Media},
	city={Wuhan},
	postcode={430079}, 
	state={Hubei},
	country={PR China}}

\affiliation[4]{organization={School of Computer Science and Technology, Huazhong University of Science and Technology},
	city={Wuhan},
	postcode={430079}, 
	state={Hubei},
	country={PR China}}

\author[1,2,3]{Kang Xue}

\ead{xuekang@mails.ccnu.edu.cn}



\author[4]{Bolong Zheng}[
orcid=0000-0001-8639-4570,
]
\ead{bolongzheng@hust.edu.cn}

\author[1,2,3]{Tingting He}[
orcid=0000-0001-7523-6550,
]
\ead{tthe@ccnu.edu.cn}
\cormark[1]

\cortext[1]{Corresponding author}

\begin{abstract}
Fine-tuning all parameters of Large Language Models (LLMs) is computationally expensive. Parameter-Efficient Fine-Tuning (PEFT) methods address this by selectively fine-tuning specific parameters. Most of the parameter efficient fine-tuning (PEFT) methods center on selecting or introducing a set of parameters to be fine-tuned. However, there are few methods that consider the impact of data samples on parameter selecting. Representative data driven methods include FISH Mask based method, which randomly selects a portion of data samples as a basis when selecting parameters. However, this random data sample selection method cannot select optimal parameters for unstable data distribution. In this work, we introduce a data-centric approach and propose the Iterative Range Decreasing (IRD) algorithm to optimize the sample-parameter pair selection in FISH Mask. IRD iteratively refines the selection by identifying subsets of samples and parameters exhibiting higher Fisher information.  We demonstrate the effectiveness and rationality of proposed strategy by conducting experiments on GLUE benchmark. Experimental results show our strategy optimizes the parameter selection and achieves preferable performance over some typical baseline methods. 
\end{abstract}

%
%
%


\maketitle

\section{Introduction}\label{}
Large language models have demonstrated fabulous capabilities across various fields through auto-regressive training on vast datasets from the internet. As generalist models, they are not optimized for any specific task during training. Therefore, supervised fine-tuning usually continues to improve the performance when facing specific problems. With the proposal of transfer learning, using a pre-trained model to fine-tune its parameters on the downstream task is becoming increasingly popular.
However, as the parameter size of the pre-trained model keeps growing (e.g., GPT~\citep{DBLP:conf/nips/BrownMRSKDNSSAA20}, 175B parameters), it becomes challenging to fine-tune all parameters of LLMs. Consequently, various methods are proposed to alleviate the GPU memory consumption and training time by freezing most of the parameters in the neural network structure and only tuning some of them ~\cite{DBLP:journals/corr/abs-2303-15647}. This method named Parameter-Efficient Fine-Tuning (PEFT) or Delta Tuning~\cite{DBLP:journals/natmi/DingQYWYSHCCCYZWLZCLTLS23}.

Several exemplary PEFT studies have been proposed, including but not limited to LoRA \cite{DBLP:conf/iclr/HuSWALWWC22}, Adapter \cite{DBLP:conf/icml/HoulsbyGJMLGAG19}, Prompt Tuning\cite{DBLP:conf/acl/LiuJFTDY022}, among others. These methodologies center on the design of new additional parameter structures or the subtly selection of specific parameters from existing networks. These existing works rarely pay attention to the impact of data on parameter selection before model training, even if it is necessary to use data for parameter selection. They just empirically select parameters or design additional structures, and then fine-tune the effectiveness of the method on the complete dataset.

Data-oriented approaches~\cite{DBLP:journals/corr/abs-2311-11696, DBLP:conf/eacl/ValipourRKG23, DBLP:journals/corr/abs-2402-04333} offer a promising direction for optimizing parameter selection in PEFT. One such method, FISH Mask~\cite{DBLP:conf/nips/SungNR21}, leverages the Fisher Information Matrix (FIM) to estimate the importance of each parameter with respect to the training data.  Specifically, FISH Mask calculates the FIM based on a randomly selected subset of training data, using the FIM's diagonal entries to rank parameter importance. This ranking then guides the selection of parameters to be fine-tuned.  However, the assumption of independent and identically distributed data and the use of random sampling for FIM calculation in FISH Mask limit its effectiveness in real-world scenarios where data distributions can be complex and non-uniform.  This raises a critical research question: How can we improve the sample selection process in FISH Mask to achieve better parameter selection and fine-tuning performance?

To address this limitation, we propose the Iterative Range Decreasing (IRD) algorithm, which iteratively refines the selection of samples and parameters based on their Fisher information.  IRD identifies a sample set with higher Fisher information content, leading to more informed parameter selection. In summary, the contributions of this work are:
\begin{itemize}
\item We investigated Parameter-Efficient Fine-Tuning (PEFT) methods based on selective strategies and observed that high-quality training samples are crucial for selecting optimal fine-tuning parameters.  Analysis of the FISH Mask method revealed that random sample selection for calculating the Fisher Information Matrix (FIM) limits its performance.

\item To address this limitation, we propose an Iterative Range Decreasing (IRD) algorithm to optimize the sample selection process.  IRD identifies a sample set with higher Fisher information, which we utilize for parameter selection during fine-tuning.

\item Extensive experiments on various representative benchmarks demonstrate the superiority of our proposed strategy.  Using only 0.2\% of model parameters for fine-tuning, our method significantly outperforms existing approaches, including widely used methods such as LoRA.
\end{itemize}

\section{Related Work}
\subsection{Parameter Efficient Fine-tuning}
With the increasing popularity of LLMs-related research, the PEFT method has developed rapidly in recent years.  As introduced in~\citet{DBLP:journals/corr/abs-2303-15647}, depending on the type of fine-tuning parameters, PEFT methods can be divided into three classes, including additive methods, selective methods, and reparametrization-based methods. Some representative additive methods include Adapter~ \cite{DBLP:conf/icml/HoulsbyGJMLGAG19}, which involves adding small, fully-connected networks after attention and feed-forward network (FFN) layers in the existing Transformer models. 

Adapters have demonstrated their effectiveness across various NLP tasks, including the GLUE benchmark, where they nearly matched the performance of fully fine-tuned BERT models. LoRA (Low-Rank Adaptation)~\cite{DBLP:conf/iclr/HuSWALWWC22} is a reparametrization-based model, which leverages the principle that neural networks can be effectively represented in low-dimensional spaces. 
This concept has received significant empirical and theoretical support~\cite{DBLP:journals/corr/abs-2003-02139,DBLP:conf/iclr/LiFLY18,DBLP:conf/icml/Arora0NZ18,DBLP:conf/icml/MalladiWYCA23}. 
LoRA utilizes a simple low-rank matrix decomposition to parameterize weight updates in neural networks, ensuring that pre-trained model weights remain frozen while only the newly added low-rank matrices are trainable.

Selective methods do not add parameters or change the model structure, which fine-tune a small part of the parameters of the model. Diff pruning ~\cite{DBLP:conf/acl/GuoRK20} and BitFit~\cite{DBLP:conf/acl/ZakenGR22} are two representative selective methods. Diff pruning learns a task-specific “diff” vector to extend the original parameters of the pretrained model. The diff vector is adaptively pruned during training to encourage sparsity. BitFit is a sparse fine-tune method, which only update the bias	of the original model during tuning. BitFit can achieve competitive  performance compared to dense full-parameter tuning.

\subsection{Fisher Information}
\label{sec:fisher}
In deep learning neural networks, the concept of Fisher information has been extensively explored and applied. The Fisher information matrix (FIM) ~\cite{fisher1922mathematical,amari1996neural} plays a crucial role in quantifying parameter importance, which is then leveraged by methods addressing the challenge of catastrophic forgetting ~\cite{french1999catastrophic,mccloskey1989catastrophic,mcclelland1995there,ratcliff1990connectionist}. While not directly mitigating forgetting, the FIM provides essential information about parameter sensitivity.  The FIM possesses three key properties ~\cite{DBLP:journals/corr/abs-1301-3584} that make it highly effective in measuring the importance of parameters within a network: (a) Relationship to the Hessian: The FIM approximates the Hessian matrix of the loss function around a minimum point. This is crucial for understanding how small changes in parameters affect the network's performance, especially near optimal points. It provides insights into the curvature of the loss landscape. (b) Computational Efficiency: The FIM can be computed using only first-order derivatives. This significantly simplifies the calculation process, making it feasible and less computationally intensive for LLMs with numerous parameters. (c) Positive Semi-Definiteness: The FIM is guaranteed to be positive semi-definite.  This ensures it describes a convex shape in the parameter space, which is important for stability during optimization.  This property ensures that parameter updates lead to predictable changes in the loss function. By quantifying the importance of parameters, methods like Elastic Weight Consolidation (EWC) ~\cite{DBLP:journals/corr/abs-1301-3584} can effectively balance learning new knowledge while preserving previously learned information. EWC leverages the FIM to identify parameters crucial for earlier tasks and applies stronger regularization to these parameters during subsequent training, preventing drastic changes and thus mitigating catastrophic forgetting.

\begin{table*}[ht]
	\small
	\centering
	\begin{tabular*}{0.9\textwidth}{c|c}
		\toprule[1.5pt]
		\textbf{Notation} & \textbf{Description}  \\
		\midrule
		$\mathcal{M}$ & The LLM to be fine-tuned. \\
		$ X_l $   &  The set of input samples in the $l$-th iteration.  \\ 
		$ Y_l = \{ y \} $   &  The set of predictable labels of sample $X$. \\ 
		$ Y^*= \{y^*\} $   &  The ground-truth label for sample $x_i$.  \\ 
		$\hat{\mathcal{I}}_\theta \in \mathbb{R}^{| \theta |}$  &  Empirical FIM that denotes the importance of each parameter.\\
		$ \Theta^*_l  $ & The set of chosen top-k parameters in the $l$-th iteration. \\ 
		$\mathbb{E}_{x \sim p(x)} \left[ f(x) \right] $   & The expected value of the function $f(x)$  \\
		& \quad when $x$ is  sampled from the distribution  $p(x)$. \\
		$p_\theta (y|x)$   & The output distribution over $y$ produced by a model with  parameter  \\ 
		& \quad vector $\theta \in \mathbb{R}^{|\theta|}$ given $x$.  \\
		$\nabla_\theta f(\theta) $   & The gradient of the function $f(\theta)$ with  respect to the parameters $\theta$.\\ 
		$S = \{ (s, \Theta^*) \}$& The set of evaluation scores with corresponding $\Theta^*$. \\
		$s={ \rm Score}(X_l, Y^*, \mathcal{M}_{\Theta^*_l}) $& Evaluation score of the model. \\
		\bottomrule[1.5pt]
	\end{tabular*}
	\caption{The frequently used notations}
	\label{tab:notations}
\end{table*}

\section{Problem Statement}
\subsection{Notations}
The frequently used notations are listed in Table~\ref{tab:notations}.

\subsection{Task Definition}
The goal of PEFT is to select a subset of parameters for fine-tuning LLMs, rather than fine-tuning all parameters.  This involves identifying a subset of existing parameters, or constructing a new set of tunable parameters (e.g., as in LoRA).  Two key factors determine the effectiveness of a PEFT method: the number of tunable parameters and the final performance.  Ideally, a PEFT method achieves comparable performance to full fine-tuning with significantly fewer parameters.  Therefore, given the same model, data, and fine-tuning objective, a smaller parameter set leading to higher performance indicates a more effective method.

\subsection{Fisher Information Matrix and FISH Mask based PEFT}

FISH Mask is a sparse selective fine-tuning method introduced by ~\citet{DBLP:conf/nips/SungNR21}. 
This method employs Fisher information to select the top-k parameters of a model. 
FIM is defined as:
\begin{equation}
	\mathcal{I}_\theta = 	\mathbb{E}_{x \sim p(x)} [ \mathbb{E}_{y \sim p_\theta(y|x)} \nabla_\theta \log p_\theta(y|x) \nabla_\theta \log p_\theta(y|x)^\textrm{T} ],
\end{equation}
where $x$ is the input, $y$ is the output, $\theta$ is the parameters of the model, $p(x)$ is the distribution of the input $x$. $\nabla_\theta$ represents the gradient of the objective function with respect to the model parameters $\theta$, computed based on the given data.
Fisher information is commonly estimated through a diagonal approximation, where gradients for all parameters are computed on several data batches: 
\begin{equation}
\hat{\mathcal{I}}_{\theta} = \frac{1}{N} \sum_{i=1}^{N} \mathbb{E}_{y \sim p_{\theta}(y|x_i)} \left[ \nabla_{\theta} \log p_{\theta}(y|x_i) \odot \nabla_{\theta} \log p_{\theta}(y|x_i) \right],
\end{equation}
where $N$ is the number of data batches. $\odot$ is Hadamard old. In supervised learning, they use ``Empirical Fisher information'' for further approximation: 
\begin{equation}
\hat{\mathcal{I}}_{\theta} = \frac{1}{N} \sum_{i=1}^{N}  \nabla_{\theta} \log p_{\theta}(y_i|x_i) \odot \nabla_{\theta} \log p_{\theta}(y_i|x_i)
\end{equation}
where $y_i$ is the ground-truth label for sample $x_i$. 
Once the top-k (empirical) Fisher information parameters are identified, only these parameters need to be optimized during fine-tuning. 
This process involves computing the (empirical) Fisher information for each parameter and creating a binary mask~\cite{DBLP:conf/emnlp/ZhaoLMJS20} based on a sparsity threshold, selectively updating only the most informative parameters in the mask. 

FISH Mask exhibits notable strengths in parameter efficiency and selective optimization. This method strategically updates a small subset of the model's parameters, rendering it particularly advantageous in scenarios where computational resources are constrained or when fine-tuning extensive models.
Additionally, by targeting the most informative parameters, FISH Mask can achieve faster convergence and shorten training duration, unlike methods that modify a broader range of parameters. 

\section{Method}

\begin{figure}[tbp]
	\centering
	\begin{minipage}[c]{1\linewidth}
		\centering
		\includegraphics[width=0.6\linewidth]{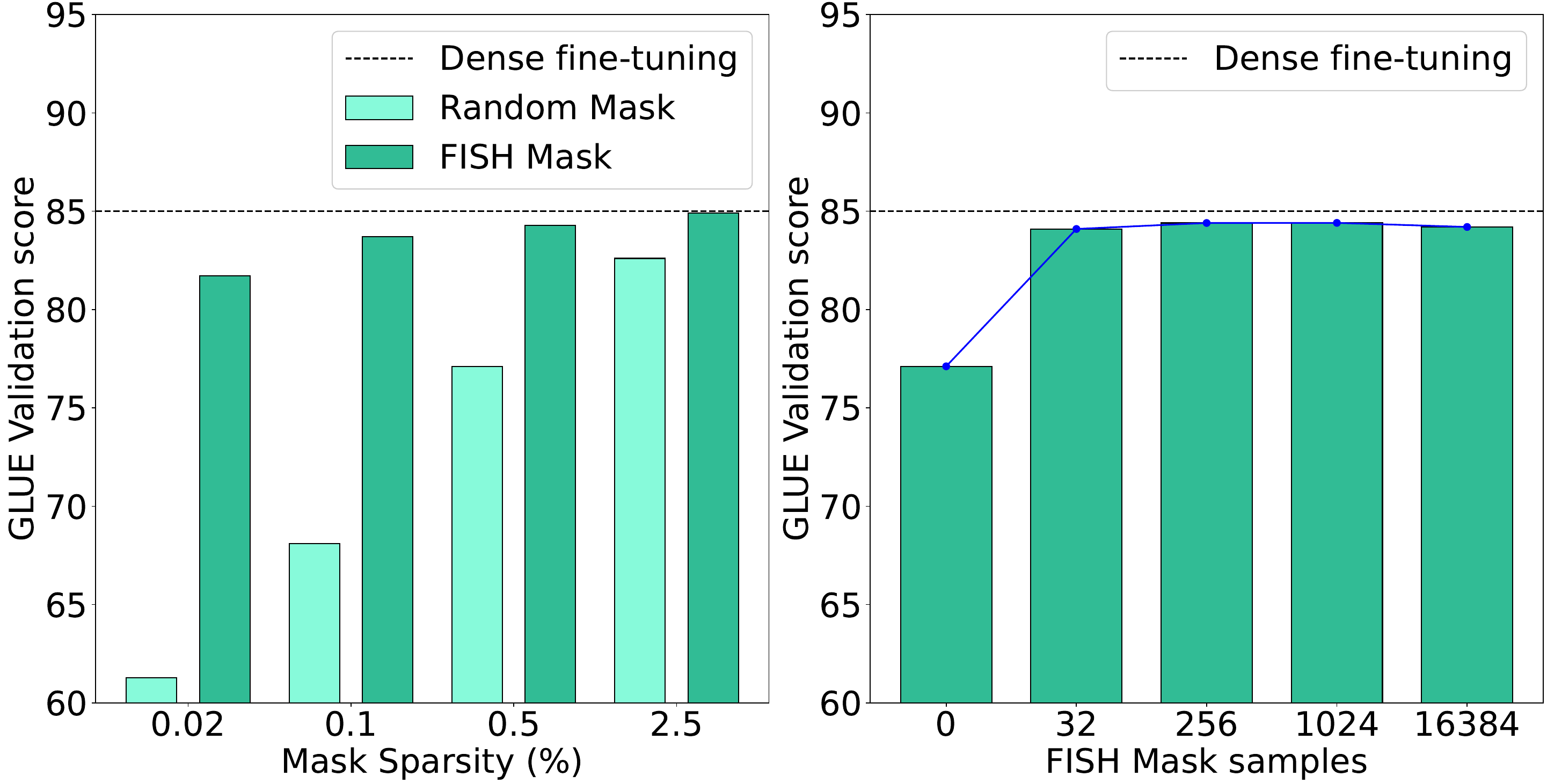}
	\end{minipage} 
	\caption{Parameter Samples Scaling Law of FISH Mask}
	\label{fig:fishsum}
\end{figure}

\subsection{Intuition}
For most of the selective-based PEFT methods, the performance of fine-tuning is related to the scale of the tuned parameters and the size of dataset, which is similar to the scaling-law theory of pretrain language model \cite{DBLP:journals/corr/abs-2001-08361}. For FISH Mask, the fine-tuning performance is affected by the parameter scale (Mask Sparsity) and the number of tuning samples, as shown in Fig.\ref{fig:fishsum}. Once the Mask Sparsity and the set of samples are fixed, FISH mask will select the optimal parameters according to the FIM of the set of samples. However, FISH Mask only focuses on the size of the sample set, and each element in the set is randomly selected. We think that this setting has drawbacks and does not consider the diversity of the samples. When the size of the sample set is determined, selecting better samples to form the set will effectively improve the effect of fine-tuning.

\subsection{Data Driven Assumption}
In  FISH Mask, only the chosen set of parameters $\theta ^*$ needs to be fine-tuned:
\begin{equation}
	\label{eq:theta}
	\Theta_0 ^* = \{ \theta_i \in \theta | \hat{\mathcal{I}}_{\theta_i} >= {\rm sort} (\hat{\mathcal{I}}_{\theta})_k  \}
\end{equation}
The setting of Eq.\ref{eq:theta} is following \citet{DBLP:conf/nips/SungNR21}, they assume that each training example $x_i$ in $\hat{ \mathcal{I}}_{\theta}$ is independent and identically distributed.
In practical datasets, maintaining a consistent data distribution proves challenging, with the presence of considerable noise data making it further complicated. Therefore, there are limitations to random select samples to calculate FIM.
In this work, we assume that the samples with larger Fisher-Information are more important in fine-tuning, which will directly affect the quality of parameter selection.

\begin{algorithm}[t]
	\small
	\caption{Iterative Range Decreasing}
	\label{algorithm:1}
	\LinesNumbered 
	\KwIn{The set of initial chosen samples $X_0$; The set of initial chosen parameters $\Theta_0$;  Ground truth of fintuned data $Y^*$; LLM to be tuned $\mathcal{M}$, $k$ is the number of initail parameters to be tuned.}
	\KwOut{$S$: Optimal parameter sets and its corresponding performance after fine-tuning.}
	$l=0$; $S = \emptyset $;
	
	$\hat{\mathcal{I}}_{\theta} = \frac{1}{N} \sum_{i=1}^{N}  \nabla_{\theta} \log p_{\theta}(y_i|x_i) \odot \nabla_{\theta} \log p_{\theta}(y_i|x_i)$
	
	$\Theta_0 ^* = \{ \theta_i \in \theta | \hat{ \mathcal{I}}_{\theta_i} >= {\rm sort} (\hat{ \mathcal{I}}_{\theta})_k  \}$;
	
	\For{$ |X_l| > 1$  and  $|\Theta_l^*| >1$}{
		$X_{l+1} = \{x_i \in X_l | \hat{ \mathcal{I}}_{x_i} >= {\rm sort}(\hat{ \mathcal{I}}_{x_i})_ \frac{|X_l|}{2} \}$;
		
		$s = {\rm Score}(X_{l+1}, Y^*,\mathcal{M}_{\Theta_l^*}) $;
		
		$S = S + \{(s, \Theta_l^*)\} $;
		
		$	\Theta_{l+1} ^* = \{ \theta_i \in \Theta_l | \hat{ \mathcal{I}}_{\theta_i} >= {\rm sort} (\hat{ \mathcal{I}}_{\theta})_\frac{|\Theta_l|}{2}  \} $
		
		$s = {\rm Score}(X_{l+1}, Y^*,\mathcal{M}_{\Theta_{l+1}^*}) $;
		
		$S = S + \{(s, \Theta_{l+1}^*)\} $;
		
		$l = l + 1$
	} 
\end{algorithm}

\subsection{Iterative Range Decreasing}
\label{sec:IRD}
We propose IRD algorithm to select better samples instead of random selecting. IRD optimizes first phase of FISH mask. The goal of IRD algorithm is to find the optimal set of parameters for fine-tuning. In each iteration, IRD reduces the scales of the set of data samples and the set of candidate parameters. The optimal parameter set obtained in each iteration is recorded. When IRD ends, we can obtain the optimal parameter set corresponding to different data scales from large to small.

As shown in Algorithm\ref{algorithm:1}, $l$ is the variable to control iteration. At the $l-$th iteration, $X_l$ is the set of data samples at and $\Theta_l^*$ is the set of parameters. $s$ is the fine-tuning performance based on the parameter set of the current iteration. The inputs of IRD algorithm are model $\mathcal{M}$ to be fine-tuned, the set of initial input samples $X_0$, and the ground-truth $Y^*$. We firstly set variable $l$ to record the iteration and $S$ to receive the output (Step 1). With these inputs, IRD firstly generates the $\Theta_0^*$, which represents the initial set of parameters with top-k Fisher information (Step 2). Then, the iterative range decreasing process is conducted. During each iteration, if the number of elements in both $X_l$ and $\Theta_l^*$ is greater than one, we alternately halve the range of both $X_l$ and $\Theta_l^*$ (Step 4 and Step 7). And the performance of $\mathcal{M}$ of specific pair of sample-parameter is calculated once the sample or parameter set changes (Step 5 and Step 8). Finally, we get a set of performance scores (calculated by evaluation metrics in Section \ref{datasets}) and a corresponding set of tuned parameters.

\subsection{Theoretical Insights into Sample Selection}

While a rigorous mathematical proof of the superiority of our proposed sample selection method is complex and depends on the specific model architecture and data distribution, we offer a simplified example to illustrate why non-random sample selection strategies can be advantageous.

\subsubsection{Simplified Linear Model}

Consider a linear model $y = \theta^T x + \epsilon$, where $x \in \mathbb{R}^d$ is the input, $y$ is the output, $\theta \in \mathbb{R}^d$ is the parameter vector, and $\epsilon \sim N(0, \sigma^2)$ is Gaussian noise.  Assuming $x \sim N(0, I)$, the Fisher information matrix simplifies to $\mathcal{I}(\theta) = \frac{1}{\sigma^2} I$.  Our goal is to select parameters that have the largest impact on the output prediction, which can be interpreted as selecting parameters along the eigenvector corresponding to the largest eigenvalue of the Fisher information matrix.

\subsubsection{Random Sample Selection}

With $n$ randomly selected samples $\{x_1, x_2, ..., x_n\}$, the empirical Fisher information matrix is $\hat{\mathcal{I}}(\theta) = \frac{1}{n\sigma^2} \sum_{i=1}^n x_i x_i^T$. Due to the randomness, $\hat{\mathcal{I}}(\theta)$ fluctuates around $\frac{1}{\sigma^2} I$, especially for small $n$.  This large variance hinders accurate estimation of the true Fisher information, making it difficult to select important parameters effectively. Due to the randomness, $\hat{\mathcal{I}}(\theta)$ fluctuates around $\frac{1}{\sigma^2} I$. This is a consequence of sampling error: the empirical Fisher information is computed from a finite sample, while the true Fisher information is based on the entire population distribution.  Specifically, $\hat{\mathcal{I}}(\theta)$ is a random matrix, as it is the average of the random matrices $x_i x_i^T$. By the law of large numbers, $\hat{\mathcal{I}}(\theta)$ converges to $\mathcal{I}(\theta)$ as $n \to \infty$. However, for finite $n$, $\hat{\mathcal{I}}(\theta)$ is merely an estimator and exhibits variability around $\mathcal{I}(\theta)$. This variability is particularly pronounced for small $n$, as a small sample set is less representative of the population and more susceptible to random fluctuations.  This increased variance in $\hat{\mathcal{I}}(\theta)$ hinders accurate estimation of the Fisher information and thus impedes effective selection of important parameters.

\subsubsection{Non-Random Sample Selection}

Suppose we strategically select samples aligned with a specific direction $v$, such that $x_i \approx \alpha_i v$, where $\alpha_i$ is a scalar.  The empirical Fisher information becomes $\hat{\mathcal{I}}(\theta) \approx \frac{1}{n\sigma^2} \sum_{i=1}^n \alpha_i^2 v v^T$.  The dominant eigenvector of this matrix aligns with $v$, indicating that we can amplify the Fisher information along $v$ by choosing samples aligned with it.  This targeted selection facilitates more efficient selection of parameters related to this direction.

\subsubsection{Advantages of Non-Random Sampling Strategies for Parameter Selection}

Random sampling yields an unbiased estimate of $\mathcal{I}(\theta)$, but with potentially high variance, especially for small $n$.  Non-random sampling such as IRD, while introducing bias, reduces the variance of the estimator. While random sampling offers an unbiased estimate of the Fisher Information matrix, $\mathcal{I}(\theta)$, non-random, or strategic, sampling presents distinct advantages for parameter selection, particularly when the objective is focused on a specific direction in parameter space. Non-random sampling allows for the targeted acquisition of information relevant to a predefined direction, $v$, in parameter space. By concentrating sampling efforts along $v$, the empirical Fisher information, $\hat{\mathcal{I}}(\theta)$, becomes more informative about the parameters that influence variations along this direction. This focused approach is particularly beneficial when prior knowledge or task-specific requirements dictate a specific direction of interest.  For example, if we know that certain parameters are more likely to impact a specific performance metric, we can tailor our sampling strategy to prioritize those parameters. Non-random sampling can significantly reduce the variance of the estimator, $\hat{\mathcal{I}}(\theta)$, compared to random sampling, especially when the sample size is limited. This reduced variance enhances the sensitivity of the estimation to changes in the parameters along the chosen direction $v$, making it easier to identify the most influential parameters. This increased sensitivity allows for more confident selection of important parameters, even with fewer samples, potentially saving computational resources.  

By prioritizing samples with high Fisher Information for subsequent iterations, IRD offers several key advantages.  This strategy focuses on the most informative samples, those that contribute most to accurately estimating the Fisher Information matrix, $\mathcal{I}(\theta)$, leading to faster convergence and more precise parameter estimates with fewer samples.  Furthermore, utilizing highly informative samples enhances the sensitivity of $\mathcal{I}(\theta)$ estimation, allowing for more accurate identification of influential parameters.  Finally, this focused approach enables efficient resource allocation by concentrating computational effort on the most informative data points, resulting in significant computational savings, especially beneficial for large datasets or complex models.

\subsection{Time and Space Complexity Analysis}
IRD is a deep learning algorithm, and its time and space complexity is difficult to accurately estimate. The time complexity of IRD depends on the number of data samples $n=|X_0|$ and the parameter scale $m=|\Theta_0^*|$ of the foundation model $\mathcal{M}$. The rough time complexity is $\log_2 nm$, where the dominant term is a fine-tuning process. The space complexity is $km$, which depends on the different optimizer. After the model and fine-tuning data set are determined, executing the IRD algorithm once can obtain multiple optimal fine-tuning parameter sets of different data scales.

\section{Experiments Setup}
All experiments can be conducted on a single GPU (RTX-4090-24GB or RTX-8000-48GB) and we mainly work on RTX-4090. Our code framework relies on the
PyTorch and Transformer framework\footnote{\url{https://pytorch.org/}, \url{https://huggingface.co/}}.

\begin{table*}[htbp]
	\centering
	\begin{tabular*}{\textwidth}{p{2.5cm}|c|ccccccccc}
		\toprule[1.5pt]
		\text{Method} & \text{Params Ratio}  & \text{QNLI}  & \text{SST-2}  & $\text{MNLI}_\text{m}$ & $\text{MNLI}_\text{mm}$ & \text{CoLA} & \text{MRPC} & \text{STS-B} & \text{RTE} & \text{QQP}  \\
		\midrule
		Dense Fine-tuning & 100\% & 93.4 & 94.9 & 87.0 & 86.1 & 61.0 & 86.6 & 86.5 & 70.9 & 80.5 \\
		\midrule
		Bit-Fit                & 0.08\% & 90.4 & 94.5 & 85.0 & 84.8 & 60.3 & 86.3 & 85.0 & 69.6 & 78.5 \\
		FISH Mask        & 0.08\% & 93.3 & 94.0 & 85.3 & 84.9 & 56.4 & 86.2 & 85.7 & 70.2 & 79.3 \\
		\midrule
		Random Mask  &  0.50\% & 89.8 & 93.4 & 83.7 & 84.0 & 43.2 & 77.8 & 87.7 & 61.3 & 77.2 \\
		Diff Pruning      & 0.50\% & 91.9 & 93.8 & 86.0 & 85.5 & 61.0 & 86.2 & 85.6 & 67.5 & 80.1 \\
		LoRA        & 0.50\% & \textbf{93.5} & 94.0 & \textbf{86.7} & 86.4 & \textbf{65.5} & \textbf{91.4} & 89.7 & \textbf{76.9} & \textbf{89.4} \\
		FISH Mask        & 0.50\% & 93.1 & \textbf{94.7} & 86.5 & 85.9 & 61.6 & 87.1 & 86.5 & 71.2 & 80.2 \\
		IRD                    & 0.50\% & 93.3& 93.7 & 86.3 & \textbf{86.6} & 62.1& 90.7 & \textbf{89.9} & 72.6 & 88.4 \\   
		\midrule
		(B)LoRA  & 0.10\%  & 90.2 & 91.1 & 81.7 & 82.3 & 59.3 & 88.1 & 84.8 & 67.5 & \textbf{86.9} \\
		(B)FISH Mask  & 0.10\%  & 89.9 & \textbf{91.3} & 79.6& 80.4 & 60.8 & 88.2 & 88.4 & \textbf{71.1} & 85.6 \\
		(B)IRD             & 0.10\%  &\textbf{ 90.4} & \textbf{91.3} &\textbf{ 82.1} & \textbf{82.4} & \textbf{61.3} & \textbf{88.6} &\textbf{ 88.8} & \textbf{71.1} & 86.1\\
		\midrule
		(B)LoRA  & 0.04\%  & 88.8 & 91.6 & 79.7 & 80.0 & 54.6 & 87.9 & 84.5 & 61.0 & 85.2 \\
		\midrule
		(B)LoRA  & 0.02\%  & 90.6 & 89.9 & 77.8 & 78.8 & 50.8 & 90.1 & 83.9 & 66.8 & 83.7 \\
		(B)FISH Mask  & 0.02\%  & \textbf{89.4} & \textbf{90.8} & 73.9 & 78.1 & 53.6 & \textbf{88} & 87.8 & 66.1 & 82.8 \\
		(B)IRD              & 0.02\%  & 89.1 & 90.7 & \textbf{79.1} & \textbf{79.7} & \textbf{54.7} & 87.9 & \textbf{87.9} & \textbf{69.3 }& \textbf{83.9 }\\
		\bottomrule[1.5pt]
	\end{tabular*}
	\caption{Performance on GLUE. (B) means the foundation model is BERT-base and the default foundation model is BERT-large. Updated Params means the ratio of fine-tuning parameters to all parameters. The Bold numbers make out the best result in the comparison with same fine-tuning ratio of parameters.}
	\label{tab:bert}
\end{table*}

\subsection{Datasets and Baselines}
\textbf{Datasets.}
\label{datasets}
We evaluate IRD on the GLUE~\cite{DBLP:conf/iclr/WangSMHLB19} benchmark dataset compared with FISH Mask method. 
GLUE is a multi-task benchmark that contains 10 tasks for LLMs evaluation. 
As task \textbf{AX} only has test set for zero-shot learning, we ignore it.  
We conduct experiments on the original data splits from GLUE.  

\textbf{CoLA}~\cite{DBLP:journals/corr/abs-1805-12471} is a 2-class classification task of English grammatical acceptability judgments.  

\textbf{SST-2}~\cite{DBLP:conf/emnlp/SocherPWCMNP13} is a 2-class classification task about human sentiment in movie reviews.  

\textbf{MRPC}~\cite{DBLP:conf/acl-iwp/DolanB05} is a 2-class classification task about senteces pair are semantically equivalent or not.  

\textbf{STS-B}~\cite{DBLP:journals/corr/abs-1708-00055} is a regression task about sentences pair similarity score from 1.0 to 5.0.  

\textbf{QQP} is a 2-class classification task about whether questions pair are semantically equivalent or not.  

\textbf{MNLI}~\cite{DBLP:conf/naacl/WilliamsNB18} is a 3-class classification task about the relationship between premise and hypothesis sentences.  

\textbf{QNLI}~\cite{DBLP:conf/emnlp/RajpurkarZLL16} is a 2-class classification task about the relationship between question and answer.  

\textbf{RTE} is a 2-class classification task from annual textual entailment challenges.  

\textbf{WNLI}~\cite{DBLP:conf/kr/LevesqueDM12} is a 2-class classification task for predicting whether the original sentence entails the sentence with the pronoun substituted.
This task only contains 635 samples for training.  

\textbf{Baselines.} Since IRD is based on the selective PEFT method. We choose some selective methods as baselines, including Bit-Fit~\cite{DBLP:conf/acl/ZakenGR22}, Diff Pruning~\cite{DBLP:conf/acl/GuoRK20}, original FISH Mask~\cite{DBLP:conf/nips/SungNR21}. Also, following the settings of~\citet{DBLP:conf/nips/SungNR21}, we introduce some simple settings as baselines, such as Dense Fine-tuning and Random Mask. Because the data set and experimental settings are exactly the same, we quoted some experimental results in~\citet{DBLP:conf/nips/SungNR21} and focused on testing IRD and the original FISH Mask method. In addition, we selected the widely used LoRA~\cite{DBLP:conf/iclr/HuSWALWWC22} method as our comparison benchmark to illustrate the effectiveness of our proposed method.

\subsection{Evaluation Metrics}
\textbf{Evaluation Metrics.} 
The metrics of CoLA task is Matthews correlation coefficient~\cite{matthews1975comparison}.
The metrics of STS-B task is Pearson and Spearman correlation coefficients.
The metrics of MRPC and QQP tasks are combined score(half the sum of the $F_1$ and Accuracy).
The metrics of other tasks are Accuracy. 
The settings of evaluation metrics is following~\citet{DBLP:conf/nips/SungNR21}.  

\begin{figure}[t]
	\centering
	\includegraphics[width=0.4\textwidth]{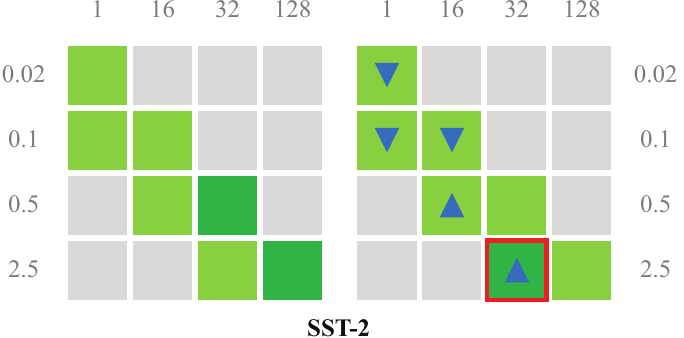}
	\caption{llustrative compared results on SST-2}
	\label{fig:caseSST-2}
\end{figure}

\begin{table}[t]
	\centering
	\tiny

	\begin{tabular}{|c|c|c|c|c|}
		\toprule[1.5pt]
		\multicolumn{5}{c}{\textbf{FISH Mask on SST-2}} \\
		\hline 
		\textbf{Mask}  & \multicolumn{4}{c|}{\textbf{FISH Mask samples}} \\ \cline{2-5}
		\textbf{Sparsity} & 1 & 16 & 32 & 128 \\
		\hline
		\multirow{2}{*}{0.02} & 0.9082- \cellcolor[rgb]{0.5725490196078431,0.8156862745098039,0.3137254901960784} & \cellcolor[rgb]{0.8509803921568627,0.8509803921568627,0.8509803921568627} &  \cellcolor[rgb]{0.8509803921568627,0.8509803921568627,0.8509803921568627} & \cellcolor[rgb]{0.8509803921568627,0.8509803921568627,0.8509803921568627} \\ 
		& 56881 \cellcolor[rgb]{0.5725490196078431,0.8156862745098039,0.3137254901960784} & \cellcolor[rgb]{0.8509803921568627,0.8509803921568627,0.8509803921568627}  & \cellcolor[rgb]{0.8509803921568627,0.8509803921568627,0.8509803921568627} & \cellcolor[rgb]{0.8509803921568627,0.8509803921568627,0.8509803921568627} \\
		\hline
		\multirow{2}{*}{0.1} & 0.9128- \cellcolor[rgb]{0.5725490196078431,0.8156862745098039,0.3137254901960784} & 0.9105- \cellcolor[rgb]{0.5725490196078431,0.8156862745098039,0.3137254901960784} &  \cellcolor[rgb]{0.8509803921568627,0.8509803921568627,0.8509803921568627} & \cellcolor[rgb]{0.8509803921568627,0.8509803921568627,0.8509803921568627} \\ 
		& 44037 \cellcolor[rgb]{0.5725490196078431,0.8156862745098039,0.3137254901960784} & 50459 \cellcolor[rgb]{0.5725490196078431,0.8156862745098039,0.3137254901960784}  & \cellcolor[rgb]{0.8509803921568627,0.8509803921568627,0.8509803921568627} & \cellcolor[rgb]{0.8509803921568627,0.8509803921568627,0.8509803921568627} \\
		\hline
		\multirow{2}{*}{0.5} & \cellcolor[rgb]{0.8509803921568627,0.8509803921568627,0.8509803921568627} & 0.9128- \cellcolor[rgb]{0.5725490196078431,0.8156862745098039,0.3137254901960784} & 0.9174- \cellcolor[rgb]{0.5725490196078431,0.8156862745098039,0.3137254901960784} & \cellcolor[rgb]{0.8509803921568627,0.8509803921568627,0.8509803921568627} \\ 
		& \cellcolor[rgb]{0.8509803921568627,0.8509803921568627,0.8509803921568627} & 44037 \cellcolor[rgb]{0.5725490196078431,0.8156862745098039,0.3137254901960784} & 31193 \cellcolor[rgb]{0.5725490196078431,0.8156862745098039,0.3137254901960784} & \cellcolor[rgb]{0.8509803921568627,0.8509803921568627,0.8509803921568627} \\
		\hline
		\multirow{2}{*}{2.5} & \cellcolor[rgb]{0.8509803921568627,0.8509803921568627,0.8509803921568627} & \cellcolor[rgb]{0.8509803921568627,0.8509803921568627,0.8509803921568627} & 0.9185- \cellcolor[rgb]{0.5725490196078431,0.8156862745098039,0.3137254901960784} & 0.9220- \cellcolor[rgb]{0,0.6901960784313725,0.3137254901960784}  \\ 
		& \cellcolor[rgb]{0.8509803921568627,0.8509803921568627,0.8509803921568627} & \cellcolor[rgb]{0.8509803921568627,0.8509803921568627,0.8509803921568627} & 77982 \cellcolor[rgb]{0.5725490196078431,0.8156862745098039,0.3137254901960784} & 18349 \cellcolor[rgb]{0,0.6901960784313725,0.3137254901960784}  \\
		
		\midrule
		
		\multicolumn{5}{c}{\textbf{IRD on SST-2}} \\
		\hline 
		\textbf{Mask}  & \multicolumn{4}{c|}{\textbf{FISH Mask samples}} \\ \cline{2-5}
		\textbf{Sparsity} & 1 & 16 & 32 & 128 \\
		\hline
		\multirow{2}{*}{0.02} & 0.9071- \cellcolor[rgb]{0.5725490196078431,0.8156862745098039,0.3137254901960784} & \cellcolor[rgb]{0.8509803921568627,0.8509803921568627,0.8509803921568627} &  \cellcolor[rgb]{0.8509803921568627,0.8509803921568627,0.8509803921568627} & \cellcolor[rgb]{0.8509803921568627,0.8509803921568627,0.8509803921568627} \\ 
		& 10092 \textcolor[rgb]{0,0.4392156862745098,0.7529411764705882}{$\downarrow$} \cellcolor[rgb]{0.5725490196078431,0.8156862745098039,0.3137254901960784} & \cellcolor[rgb]{0.8509803921568627,0.8509803921568627,0.8509803921568627}  & \cellcolor[rgb]{0.8509803921568627,0.8509803921568627,0.8509803921568627} & \cellcolor[rgb]{0.8509803921568627,0.8509803921568627,0.8509803921568627} \\
		\hline
		\multirow{2}{*}{0.1} & 0.9128- \cellcolor[rgb]{0.5725490196078431,0.8156862745098039,0.3137254901960784} & 0.9116- \cellcolor[rgb]{0.5725490196078431,0.8156862745098039,0.3137254901960784} &  \cellcolor[rgb]{0.8509803921568627,0.8509803921568627,0.8509803921568627} & \cellcolor[rgb]{0.8509803921568627,0.8509803921568627,0.8509803921568627} \\ 
		& 44037 \cellcolor[rgb]{0.5725490196078431,0.8156862745098039,0.3137254901960784} & 97248 \textcolor[rgb]{0,0.4392156862745098,0.7529411764705882}{$\uparrow$} \cellcolor[rgb]{0.5725490196078431,0.8156862745098039,0.3137254901960784}  & \cellcolor[rgb]{0.8509803921568627,0.8509803921568627,0.8509803921568627} & \cellcolor[rgb]{0.8509803921568627,0.8509803921568627,0.8509803921568627} \\
		\hline
		\multirow{2}{*}{0.5} & \cellcolor[rgb]{0.8509803921568627,0.8509803921568627,0.8509803921568627} & 0.9174- \cellcolor[rgb]{0.5725490196078431,0.8156862745098039,0.3137254901960784} & 0.9162- \cellcolor[rgb]{0.5725490196078431,0.8156862745098039,0.3137254901960784} & \cellcolor[rgb]{0.8509803921568627,0.8509803921568627,0.8509803921568627} \\ 
		& \cellcolor[rgb]{0.8509803921568627,0.8509803921568627,0.8509803921568627} & 31193 \textcolor[rgb]{0,0.4392156862745098,0.7529411764705882}{$\uparrow$} \cellcolor[rgb]{0.5725490196078431,0.8156862745098039,0.3137254901960784} & 84404 \textcolor[rgb]{0,0.4392156862745098,0.7529411764705882}{$\downarrow$} \cellcolor[rgb]{0.5725490196078431,0.8156862745098039,0.3137254901960784} & \cellcolor[rgb]{0.8509803921568627,0.8509803921568627,0.8509803921568627} \\
		\hline
		\multirow{2}{*}{2.5} & \cellcolor[rgb]{0.8509803921568627,0.8509803921568627,0.8509803921568627} & \cellcolor[rgb]{0.8509803921568627,0.8509803921568627,0.8509803921568627} & 0.9162- \cellcolor[rgb]{0.5725490196078431,0.8156862745098039,0.3137254901960784} & 0.9220- \cellcolor[rgb]{0,0.6901960784313725,0.3137254901960784}  \\ 
		& \cellcolor[rgb]{0.8509803921568627,0.8509803921568627,0.8509803921568627} & \cellcolor[rgb]{0.8509803921568627,0.8509803921568627,0.8509803921568627} & 84404 \textcolor[rgb]{0,0.4392156862745098,0.7529411764705882}{$\downarrow$} \cellcolor[rgb]{0.5725490196078431,0.8156862745098039,0.3137254901960784} & 18349 \cellcolor[rgb]{0,0.6901960784313725,0.3137254901960784}  \\
		\bottomrule[1.5pt]
	\end{tabular}
	
	\caption{Accurate compared results on SST-2 by BERT (In order to display more accurate values, the numbers in the table will be wrapped.)}
	\label{tab:value_SST2}
\end{table}

\begin{figure*}[tb]
	\centering
	\includegraphics[width=0.9\textwidth]{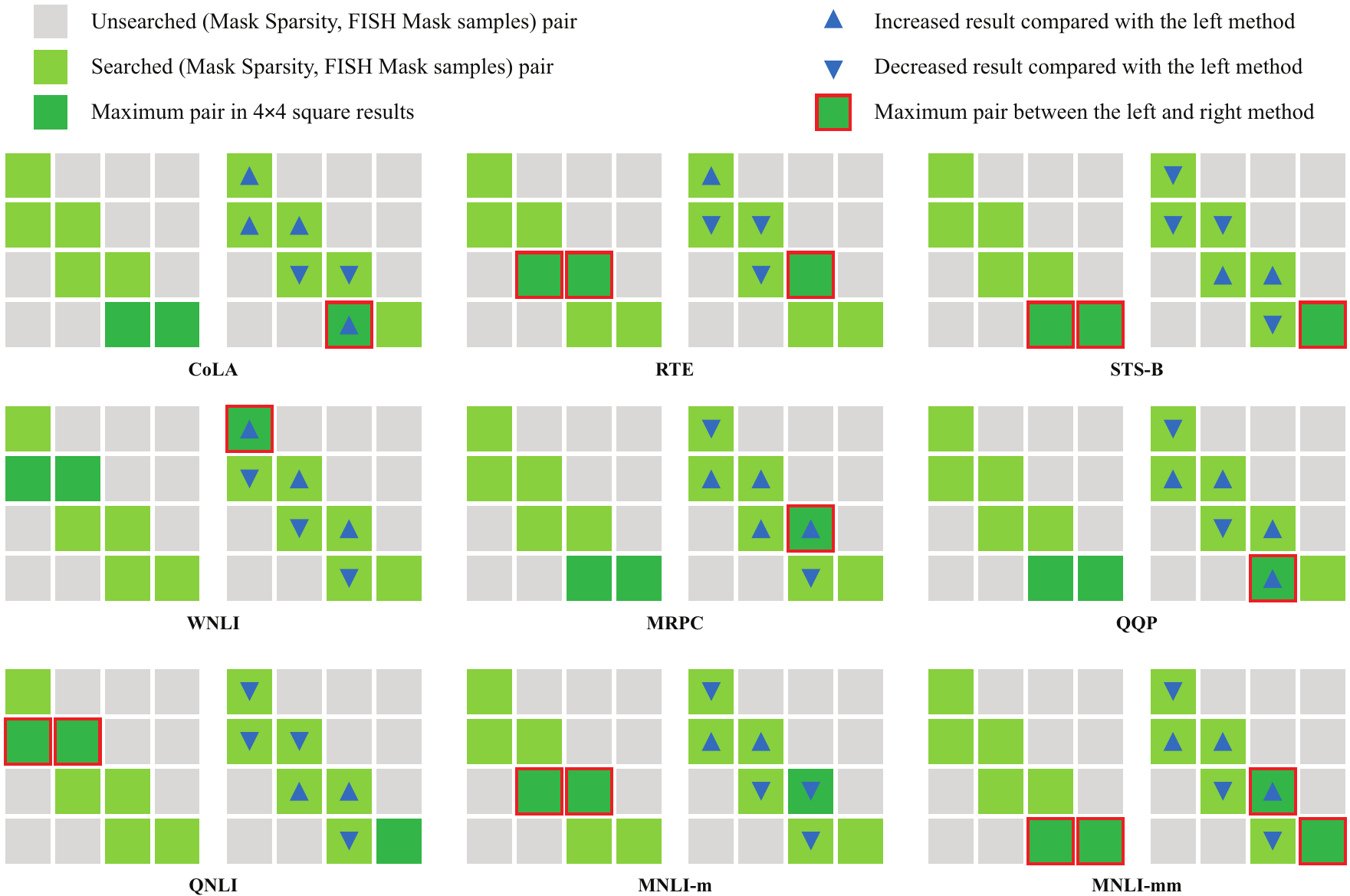}
	\caption{Comparison results between FISH Mask method and IRD-based optimization method on BERT-base}
	\label{fig:result-bert}
\end{figure*}

\subsection{Implementation Details}
\label{sec:details}
We only use the validation set results as the final results because of the limitation of 2-time uploads to the GLUE official website each day.
We follow the same parameter settings for the classification task as BERT~\cite{DBLP:conf/naacl/DevlinCLT19}.
Apart from the BERT model, we also compare IRD with FISH Mask in the auto-regressive model like GPT-2~\cite{radford2019language}.
We use early stopping to prevent overfit and underfit.
The prompt template for GPT-2 is following the Table 1 in~\citet{DBLP:journals/corr/abs-2302-10198} . 
\begin{figure*}[htb]
	\centering
	\includegraphics[width=0.9\textwidth]{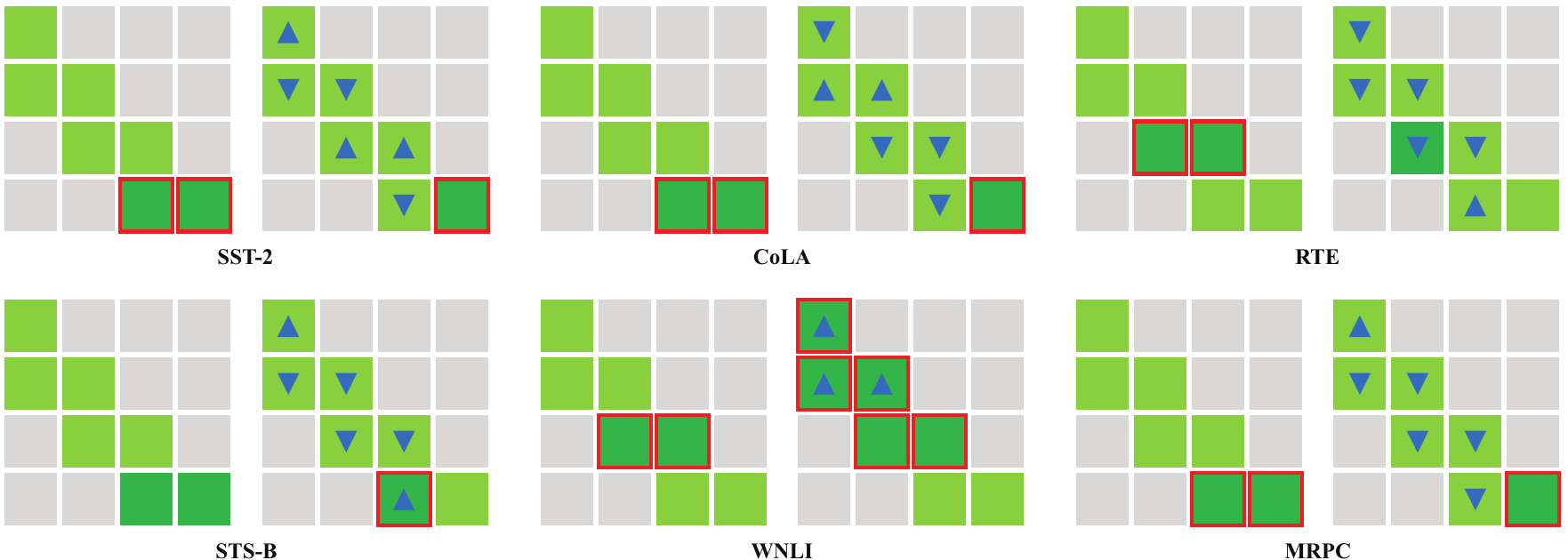}
	\caption{Comparison results between FISH Mask method and IRD-based optimization model on GPT-2}
	\label{fig:result-GPT}
\end{figure*}

We set different $max\_seq\_length$ for each task in GPT-2, shown in Table~\ref{tab:detailGPT2}, because of the different maximum sample lengths in different tasks.  
\begin{table}[ht]
	\centering
	\begin{tabular}{c|c|c|c|c|c}
		\toprule[1.5pt]
		\textbf{Task} & \textbf{max\_seq\_length} & \textbf{Task} & \textbf{max\_seq\_length}& \textbf{Task} & \textbf{max\_seq\_length} \\
			\midrule
			SST-2 & 128 & CoLA & 128  & RTE & 384 \\
			STS-B & 256 & WNLI & 128 & MRPC & 128 \\
			QQP & 512 &  QNLI & 512 & MNLI & 512 \\
			\bottomrule[1.5pt]
		\end{tabular}
		\caption{The different setting details in different task for GPT-2.}
		\label{tab:detailGPT2}
	\end{table}

For the early stopping mechanism, we uses a patience parameter of 10 epochs and a threshold value of 0.3 to determine stopping criteria. 
Training is halted if there's no improvement exceeding the patience parameter over 10 consecutive epochs and if the metric value exceeds the threshold value of 0.3. 
The $learning\_rate$ is 5e-5 in most of the tasks, except WNLI is 5e-6.
If the training context only has one sentence, we use "[CLS] sentence \uppercase\expandafter{\romannumeral1}". Otherwise we use "[CLS] sentence \uppercase\expandafter{\romannumeral1} [SEP] sentence \uppercase\expandafter{\romannumeral2}" to concatenate each context.
The $batch\_size$ is set to 32. The loss function is cross-entropy and the optimizer is Adam~\cite{DBLP:journals/corr/KingmaB14}. 

\begin{figure*}[tb]
	\centering
	\includegraphics[width=0.9\textwidth]{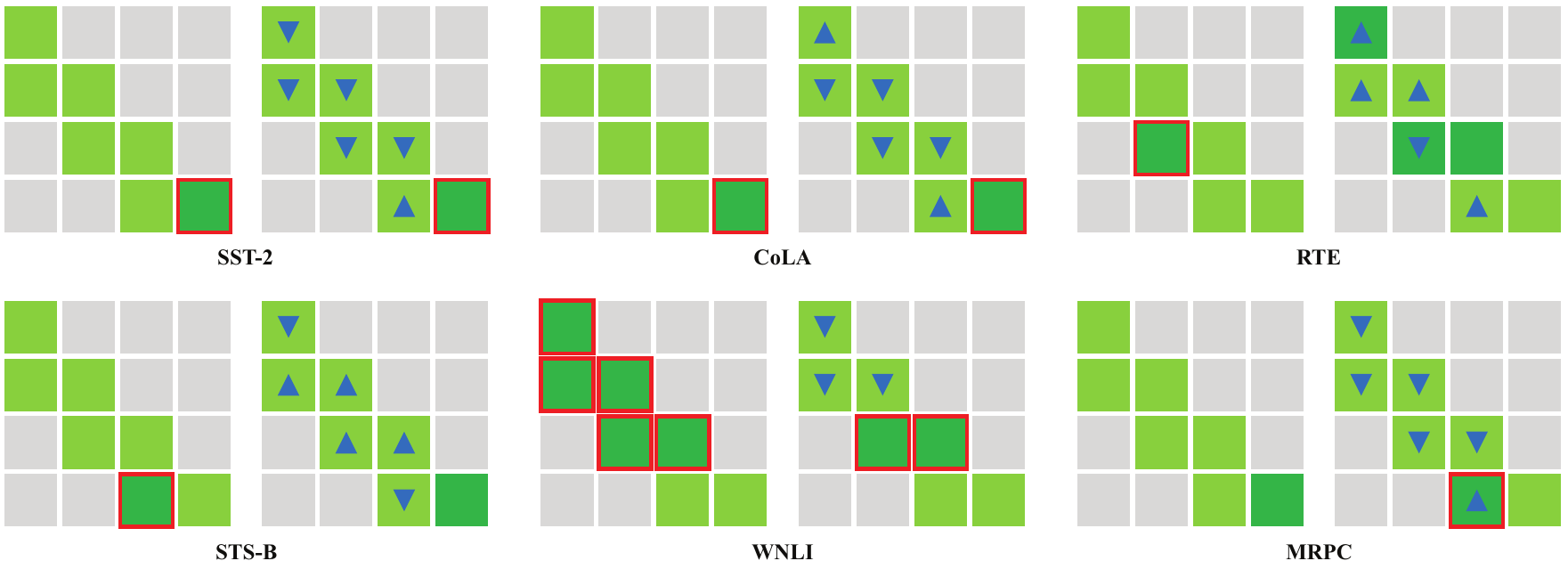}
	\caption{Contrastive Study}
	\label{fig:result-Contrastive}
\end{figure*}

\subsection{Experimental Results}
We conduct extensive experiments on the GLUE benchmark to verify the effectiveness of the IRD algorithm. Restricted by page limitation, we use pictures instead of specific values, and distinguish the values through color shades and different marks. We will use an example in Section \ref{sec:example} to show how Table \ref{tab:value_SST2} is replaced by Fig.\ref{fig:caseSST-2}, and follow this method to draw all experimental results. The remaining numerical values corresponding to illustrated experimental results will be placed in the Appendix.

\subsubsection{Case on SST-2 task: An Example}
\label{sec:example}
In Fig.~\ref{fig:caseSST-2}, we present a comparative visualization of the FISH Mask method versus IRD, which is tested on the SST-2 task using the BERT model.  Result of each GLUE task is structured into two 4 by 4 matrices:  the one on the left displays the results of the FISH Mask,  while the one on the right showcases the results from IRD.  Each block within these matrices corresponds to the fine-tuned score of a unique combination of Mask Sparsity and FISH Mask samples,  as detailed in Table~\ref{tab:value_SST2}.  The varying shades of green color within the cells signify the performance metrics,  with darker hues denoting the highest values achieved in each matrix. Besides, the unexplored combinations are indicated by grey blocks. 

The lower right corner of the result table and the 4 by 4 matrix both correspond to the original sample-parameter set pair, and are reduced in size in turn through the IRD algorithm. The sample set size decreases from right to left, and the parameter size decreases from bottom to top. Therefore, using the IRD algorithm to search from the lower right corner to the upper left corner will produce a green ladder-like effect that ascends from the lower right to the upper left corner.

Notably, triangles pointing upwards in the right matrix denote outcomes that surpass their counterparts in the left matrix, a relationship mirrored by the arrow indicators in Table\ref{tab:value_SST2}, with downward pointing triangles indicating the opposite.  Additionally, the dark green blocks with red borders highlight the overall maximum results across the entire two 4 by 4 matrices layout.  Opting for graphical representation over tabular data,  we prioritize clarity and efficient space usage in conveying our experimental findings. In experiment, the initial FISH mask sample is 128 and the initial mask sparsity is 2.5 which is set according to the parameter samples scaling law of FISH Mask shown in Fig.\ref{fig:fishsum}.

\subsubsection{Results with BERT on GLUE}
Table~\ref{tab:bert} shows the detailed performance of IRD and baseline methods on the GLUE benchmark. We conducted 8 experiments based on BERT-large foundation model. With the scale of 0.5\% parameters, LoRA achieves the best performance on 6 out of 9 GLUE subtasks. Besides, under the BERT-base foundation model with 0.1\% parameters to be fine-tuned,  IRD achieves the best performance on 8 out of 9 GLUE subtasks. On a smaller parameter scale (0.02\%), IRD can achieve similar performance to LoRA with only half the amount of fine-tuning parameters, and is better than FISH Mask. The parameter scale of LoRA is affected by rank and cannot be linearly adjusted to 0.02\%.

Fig.\ref{fig:result-bert} shows the detailed results of experiments conducted on the GLUE benchmark under BERT-base model,  
excluding SST-2.  Across these tasks,  IRD shows better results  (more increased arrows than decreased arrows) than the FISH Mask method in CoLA, RTE, STS-B, and MRPC, while it underperforms in WNLI.  In QQP, QNLI, MNLI-m, and MNLI-mm, IRD, and FISH Mask call a draw. In conclusion, experimental results prove that the IRD method is better than FISH Mask. There are far more upward arrows than downward arrows (30>22, indicating that IRD is better), which proves the reasonableness and superiority of IRD.

\subsubsection{Results with GPT-2 on GLUE}
To demonstrate the impact of IRD on the transformer-based decoder-only models,  we do experiments on the GPT-2 pre-trained model.  Fig.~\ref{fig:result-GPT} reveals the results of experiments conducted on the GLUE benchmark, using the GPT-2 model.  As we have set different $max\_seq\_length$ for each task in GPT-2 that are displayed in Section~\ref{sec:details},  we only run the experiments on six tasks from the GLUE benchmark due to constraints in training time and GPU usage.  In SST-2, STS-B, WNLI, and MRPC, IRD achieves better performance than FISH Mask. This result effectively proves the generalizability of IRD under different foundation models.

\begin{figure*}[tb]
	\centering
	\includegraphics[width=0.9\textwidth]{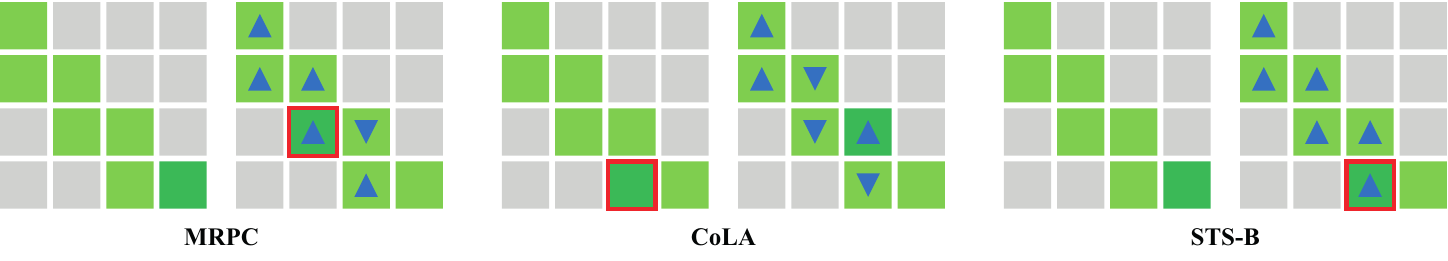}
	\caption{Results with LLaMA}
	\label{fig:result-LLaMA}
\end{figure*}

\subsubsection{Results with LLaMA on GLUE}
To further investigate the efficacy of IRD on LLaMA3.2, we conduct experiments using the LLaMA pre-trained model. Fig.~\ref{fig:result-LLaMA} presents the results on the GLUE benchmark.  Similar to the GPT-2 experiments, we adjust the $max_seq_length$ parameter for each task. Due to computational constraints, we focus on four tasks from the GLUE benchmark.  On the MRTC, COLA, and STS-B tasks, IRD demonstrates superior performance. This suggests that the benefits of IRD might be more pronounced in larger models like LLaMA. This further supports the effectiveness and generalizability of the IRD strategy across different foundation model architectures and scales. We used the full parameter search of the LLaMA 1B model, and the experimental results verified the effectiveness of the IRD method. On larger LLMs, we can use IRD for several deeper layers of the model to improve efficiency. Many existing studies have also confirmed that the last few layers of the model contain more common sense knowledge information.~\cite{DBLP:conf/emnlp/WangYXQD00GJX0C24} 

\subsubsection{Contrastive Study}
To demonstrate the enhancement of IRD over the FISH Mask method,  we switch the sample choosing step (Step 4) and parameter choosing step(Step 7) of Algorithm~\ref{algorithm:1} to a contrastive way, aiming for reverse results.  In Algorithm~\ref{algorithm:1}, Step 4 represents selecting the larger half of elements from set $X_l$ to form a new set.  Here, we modify it to select the larger half of elements from set $X_l$, but this time from the latter half (Shown in Eq.~\ref{eq5}).
In Eq.~\ref{eq6}, we make the same operation for the set of $\Theta_l^*$.
\begin{equation}
	X_{l+1} = \{x_i \in X_l | \hat{ \mathcal{I}}_{x_i} < {\rm sort}(\hat{ \mathcal{I}}_{x_i})_ \frac{|X_l|}{2} \}
	\label{eq5}
\end{equation}
\begin{equation}
	\Theta_{l+1} ^* = \{ \theta_i \in \Theta_l | \hat{ \mathcal{I}}_{\theta_i} < {\rm sort} (\hat{ \mathcal{I}}_{\theta})_\frac{|\Theta_l|}{2}  \}
	\label{eq6}
\end{equation}
Fig.\ref{fig:result-Contrastive} reveals the results of experiments conducted on the GLUE dataset, 
using the GPT-2 model. As shown in Fig.\ref{fig:result-Contrastive}, the result on each GLUE task is structured into two 4 by 4 matrices: 
the one on the left displays the results of IRD, 
while the one on the right showcases the results from the contrastive method.  

In non-tie results, IRD is better than FISH mask. We can see the obvious advantages of IRD from Fig.4. We set IRD to search in the opposite direction of optimization, that is, search for worse values in each iteration, and put the results on the right, and IRD on the left for comparison in Fig.4. Experimental results prove that the IRD method is better than reverse IRD on four subtasks. There are far more downward arrows than upward arrows (20>12, indicating that reverse IRD reduces performance), which proves the reasonableness and superiority of IRD search.

Across these six tasks, 
the modified algorithm shows more improvements (more increased arrows than decreased arrows) than IRD in RTE, STS-B, 
while it underperforms in SST-2, CoLA, WNLI, and MRPC. 
Regarding the overall results in the two matrices, 
the modified algorithm achieves the highest results in MRPC. 
IRD outperforms the modified algorithm in RTE and STS-B, 
while both methods tie for the highest results in SST-2, CoLA, and WNLI. 

\subsection{Compared with LoRA}
As a PEFT method, LoRA\cite{DBLP:conf/iclr/HuSWALWWC22} has attracted a huge amount of attention for its low computing cost and satisfactory performance. LoRA and methods based on it \cite{DBLP:conf/nips/DettmersPHZ23} became a widely compared benchmark. In order to verify the effectiveness of the IRD algorithm, we compared it with LoRA on a variety of different parameter scales. The experimental results are shown in Table \ref{tab:bert}. It shows that LoRA achieves better performance when using 0.5\% parameter of BERT. Meanwhile, IRD shows better performance when using 0.1\% parameter of foundation model. Since the parameter scale of LoRA depends on the rank of additive matrix, we set the rank as 1 and get the least parameter scale of LoRA on BERT base without changing the other settings (number of layers or number of matrix). LoRA only achieved similar results to IRD with twice the parameter scale (0.04\% versus 0.02\%). It reveals that the fine-tuning effect is not as good as IRD when the parameter scale is small.

Based on the structure of the LoRA model, it has two disadvantages compared to IRD: (1) Limited by the rank attribute of LoRA, it is impossible to finely adjust the parameter scale linearly. (2) The fine-tuning performance at a smaller parameter scale is not as good as that of IRD. (3) Compared with the selective-based PEFT method, LoRA requires additional parameter storage space. At the same time, we also believe that LoRA does have better computing efficiency and lower computational complexity.
\subsection{Analysis}
In this work, we design different types of experiments to evaluate IRD algorithm.  The chosen GLUE benchmark can fully verify the generalization of the model and facilitate comparison with other methods. By thoroughly compared with the mainstream selective-based PEFT method without changing the dataset is sufficient to demonstrate the effectiveness of our method. Besides, comparing with FISH-Mask under different foundation models further demonstrates our method's efficacy.
A contrastive study shows our method has improved on more corresponding squares and also achieved optimal values on more tasks. These results demonstrate IRD is an effective optimization algorithm because the reverse settings get worse results. 

It is worth noting that when the optimal result (red border square) appears in the lower right corner of the 4 by 4 matrix, it means that the model has achieved the best result on the initial sample-parameter size. In this case, the FISH-Mask method and the IRD algorithm get in a draw. This is because the optimal sample-parameter pair appears outside the initial range, and the IRD algorithm cannot achieve better results by continuing to decrease the sample-parameter range.

\section{Conclusion}
In this paper, we adopt a data-oriented perspective to optimize PEFT method before training. Based on this methodology, we propose IRD algorithm to optimize FISH Mask based method. Besides, we designed and conducted experiments to verify the effectiveness of the proposed algorithm, and the experimental results also verified our methodology.  We hope our efforts can inspire research in related fields, directing more attention towards data-driven PEFT methods. In future work, we will explore a set of general data-oriented PEFT optimization algorithms instead of just optimizing a certain model. In addition, we will try and explore the relationship between PEFT method data and parameters more deeply.

\section{Acknowledgement}
This work is partially supported by Hubei Provincial Natural Science Foundation (No.2023AFB487), China Postdoctoral Science Foundation (No.2023M731253), the Key Research and Development Program of Hubei Province (2020BAB017)

\printcredits

\bibliographystyle{cas-model2-names}

\bibliography{refs}



\end{document}